\pdfoutput=1

\documentclass[11pt]{article}

\usepackage[final]{coling}

\usepackage{times}
\usepackage{latexsym}
\usepackage{amsmath}

\usepackage{newfloat}
\usepackage{listings}
\usepackage{colortbl}
\definecolor{Gray}{gray}{0.9}

\usepackage{pifont}
\usepackage{fontawesome}

\newcommand{\cmark}{\ding{51}}%
\newcommand{\xmark}{\ding{55}}%

\usepackage[T1]{fontenc}

\usepackage[utf8]{inputenc}

\usepackage{microtype}

\usepackage{inconsolata}

\usepackage{graphicx}

\title{Acquired \texttt{TASTE}: \\Multimodal Stance Detection with Textual and Structural Embeddings}

\author{Guy Barel \\
  Ben Gurion University \\
  \texttt{guybare@post.bgu.ac.il} \\\And
  Oren Tsur \\
  Ben Gurion University \\
  \texttt{orentsur@bgu.ac.il} \\\And
  Dan Vilenchik \\
  Ben Gurion University \\
  \texttt{vilenchi@bgu.ac.il} \\}

\begin{document}
\maketitle
\begin{abstract}
Stance detection plays a pivotal role in enabling an extensive range of downstream applications, from discourse parsing to tracing the spread of fake news and the denial of scientific facts. While most stance classification models rely on the textual representation of the utterance in question, prior work has demonstrated the importance of the conversational context in stance detection.  
In this work, we introduce TASTE -- a multimodal architecture for stance detection that harmoniously fuses Transformer-based content embedding with unsupervised structural embedding. Through the fine-tuning of a pre-trained transformer and the amalgamation with social embedding via a Gated Residual Network (GRN) layer, our model adeptly captures the complex interplay between content and conversational structure in determining stance. TASTE achieves state-of-the-art results on common benchmarks, significantly outperforming an array of strong baselines.
Comparative evaluations underscore the benefits of social grounding -- emphasizing the criticality of concurrently harnessing both content and structure for enhanced stance detection.
\end{abstract}

\section{Introduction}

\label{sec:intro}
Stance detection is the task in which the attitude of a speaker towards a target (individual, idea, proposition) is identified. A typical example provided by \citet{somasundaran2010recognizing}: Given an utterance U (\emph{``Government is a disease pretending to be its own cure''}) and a target T (\texttt{universal health care}), we wish to determine the speaker's stance, as reflected in U toward the target T. Notice that the target is not explicitly mentioned by the speaker. A shared task with a similar setting was introduced in SemEval by \citet{mohammad2016semeval}.

However, opinions are not expressed in a vacuum but rather in a conversational context. The speaker's stance may emerge as the conversation evolves. This idea echoes Goffman's classic assertion in Response Cries \cite{goffman1978response}: ``Utterances are not housed in paragraphs but in turns at talk, occasions implying a temporary taking of the floor as well as alternation of takers''.

Indeed, some works look beyond the utterance level and address its context -- whether the speaker level or the conversational level \cite{walker2012stance,sridhar-etal-2015-joint,johnson2016identifying,joseph2017constance,li2018structured,pujari2021understanding}, among others. 
Stance detection based on the conversation structure alone was recently demonstrated by \citet{pick2022stem}.

Focusing on conversational stance, we are interested in detecting stance in a multi-participant conversation, both on the utterance and speaker level. To this end, we propose TASTE -- a multimodal architecture combining Textual And STructural Embeddings. Multi-participant conversations unfold in tree structures, which can be converted into speakers' interactions graphs (see details in Section \ref{subsec:model}). TASTE uses LLMs to represent utterances; the max-cut graph optimization problem is used to derive contextual node (speaker) embeddings from the interaction graph. The embeddings of the two modalities are fused through a GRN layer. We considered several other text and node representations as well as different ways to combine them in a principled way.

Our architecture has two key benefits:
First, it consistently outperforms an array of strong baselines, including the state-of-the-art, across topics in commonly used datasets. 
Second, through ablation and careful analysis, we get a glimpse of the interdependence of text and conversation structure. 

We find that the heavy lifting is achieved by the conversation structure. Teasing text and structure modalities apart, we find that, in most cases,  using the structure alone outperforms text-based models. Combining the textual modality with the structural further adds, on average, 12\% to the accuracy. 
This result should not be read as ``structure is more important than text''. Rather, we maintain that the stance signal is encoded in a structure with a higher signal-to-noise ratio compared to text. This makes sense as texts are produced within a conversational context by subjective individuals. We further discuss this idea in Section \ref{sec:discussion} as it goes back to the sociolinguistics concepts of face \cite{goffman1955face}, conversational norms \cite{grice1975logic}, and communication grounding \cite{clark1991grounding}. 

\begin{figure*}[ht!]
    \centering
    \includegraphics[width=1\linewidth]{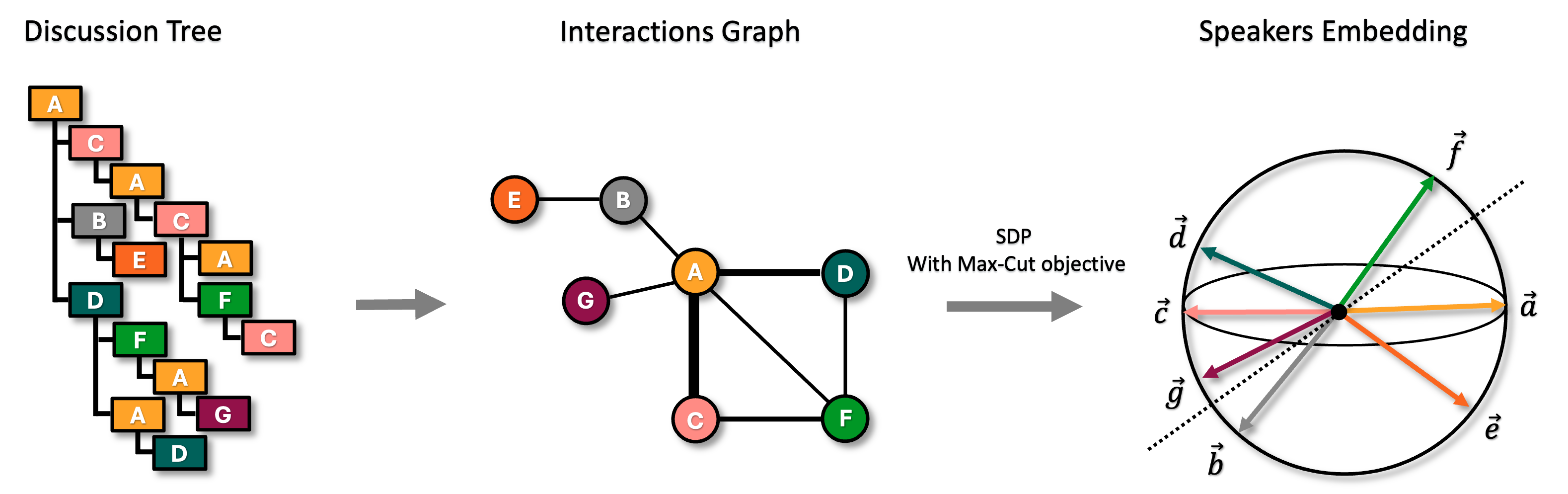}
    \caption{Illustration of a discussion tree, its corresponding speakers interactions graph, and the derived speakers embedding using the max-cut SDP. Node colors correspond to speakers. Tree nodes represent utterances; Graph nodes represent speakers. Edge width corresponds to the number of interactions. Speakers' embeddings lie on an $n$-dimensional sphere and are rounded to discrete values by projection onto a random hyperplane (dashed line).}
    \label{fig:tree2graph}
\end{figure*}

\section{Related Work}
\label{sec:rw}
\paragraph{Stance Detection} Early work tackled stance detection by identifying specific lexical forms. \citet{biber1988adverbial} consider six adverbial categories that mark the speaker's stance. Lexicons of word categories reflecting psychological states such as LIWC \cite{tausczik2010psychological} and MPQA \cite{wilson2005recognizing} were used effectively by \cite{somasundaran2010recognizing,murakami-raymond-2010-support,yin-etal-2012-unifying,walker2012stance,wang-cardie-2014-improving,elfardy-diab-2016-cu} among others. 

Recent works fine-tune LLMs \cite{samih2021few,kawintiranon2021knowledge}, or employ zero-shot \cite{allaway2020zero,liang2022zero,allaway2023zero,weinzierl2024tree} and few-shot frameworks \cite{liu2021enhancing,liu2022target,wen2023zero,khiabani2024fewshot}. 
These works use the textual content of an utterance as the sole or primary signal. 

Speaker (user) and social attributes are extracted and used to enrich textual attributes in training regression and SVM models to predict utterance-level stance \cite{aldayel2019your,lynn2019tweet}, while \citet{murakami-raymond-2010-support,walker2012corpus,yin-etal-2012-unifying} use the conversation structure with textual markers. Branch-Bert \cite{li2023improved} uses part of the conversation structure, i.e. a branch in the conversation tree in order to provide their stance classifier with some contextual information. While it is shown that the naive turn-taking structure that is captured in a branch does not fair well with graph-based embeddings \cite{pick2022stem}, a turn-taking modeling combined with a LLM achieved improved results comparing to text-only models \cite{wen2024transitive}. 

The relationship between the post and the user level is addressed by \citet{sridhar-etal-2015-joint,benton2018using,li2018structured,conforti2020will}.
A hierarchical model combining text and user representation is proposed by \citet{porco2020predicting}. 
\citet{pick2022stem} and \citet{zhang2023doubleh} demonstrate that structure alone can achieve competitive results on both the utterance and the speaker levels.

These works, using or combining the different modalities, serve as our baseline models (see Section \ref{sec:exp}) and as the reference point for our analysis and discussion (Sections \ref{sec:res}\&\ref{sec:discussion}).

\paragraph{Multimodality}
Modalities indicate the different \emph{mediums} that information is conveyed through: text (in natural language), code (formal language), audio, visual, etc. Using more than a single modality achieves an improvement in the performance of many ML systems on multiple tasks. In multimodal learning, the representations of the different modalities are used explicitly in the training phase \cite{ramachandram2017deep,multimodalTransformersSurvey2023}, rather than just defining a feature vector, e.g., \cite{aldayel2019your,lynn2019tweet}.

Naturally, the common modalities that are used are those with mature uni-modal frameworks such as Transformers and CNNs for text and image processing and generation \cite{mao2016training,yu2017multi,sharma2018conceptual,ramesh2021zero,ding2022cogview2,rombach2022high,ramesh2022hierarchical,kwak2023multimodal}; modalities that are closely related such as conversational audio with its transcribed text, e.g., \cite{lai2019detecting,yao2020morse,yao-mihalcea-2022-modality}, or code segments coupled with documentation, e.g., \cite{kwak2023multimodal} and commercial products like Github's Copilot, OpenAI's Codex, and Google's Gemini. 

Two of the works mentioned above explicitly address multi-modality in the context of stance detection. However, we note that the second modality used by \cite{weinzierl2024tree} is the visual (image) modality; Multimodality in \citet{khiabani2024fewshot} refers to the use the different modalities independently to train a number of independent classifiers. 

In this paper, we apply the multimodal framework over two modalities: text and social context.

\section{Methodology}
\label{sec:method}

\subsection{Task Definition}
Given a set of authors $A$ participating in a discussion, the set $U^a = \{u^a_{1}, u^a_{2}, ..., u^a_{n}\}$ denotes the utterances produced by $a \in A$ throughout the discussion, where each utterance $u^a_j$ and user $a$ have a label $y_{j,a},y_a \in \{+,-\}$ (\emph{pro}, \emph{con}). 

In this work, we address two distinct stance prediction tasks: utterance-level and author-level. At the utterance level, the classification task is straightforward -- learning a classifier $\sigma(u^a_j,\cdot)=\hat{y}_{a,j}$ that minimizes some loss function $\ell(y_{j,a},\hat{y}_{a,j})$. The dot in $\sigma$ represents additional input that the classifier may take besides the utterance. In our case, this extra input is the speakers' interactions graph (see Figure \ref{fig:tree2graph}). In other cases that can be meta-data like the speaker's age, gender, etc.

At the author/speaker level, we assume stance labels are assigned to speakers. This assumption is valid if a speaker holds a pre-formed and stable stance throughout the discussion. This is inherently the case in debate-like discussions like those in our datasets. 
The task, in this case, is to train a classifier $\sigma(U^a,\cdot)=\hat{y}_a$,  in which all the utterances of a user are fed to the classifier, plus additional data (the graph in our case).

\subsection{Model Architecture and Components}
\label{subsec:model}

\begin{figure}[ht!]
    \centering
    \includegraphics[width=0.95\columnwidth, height=14.7cm]{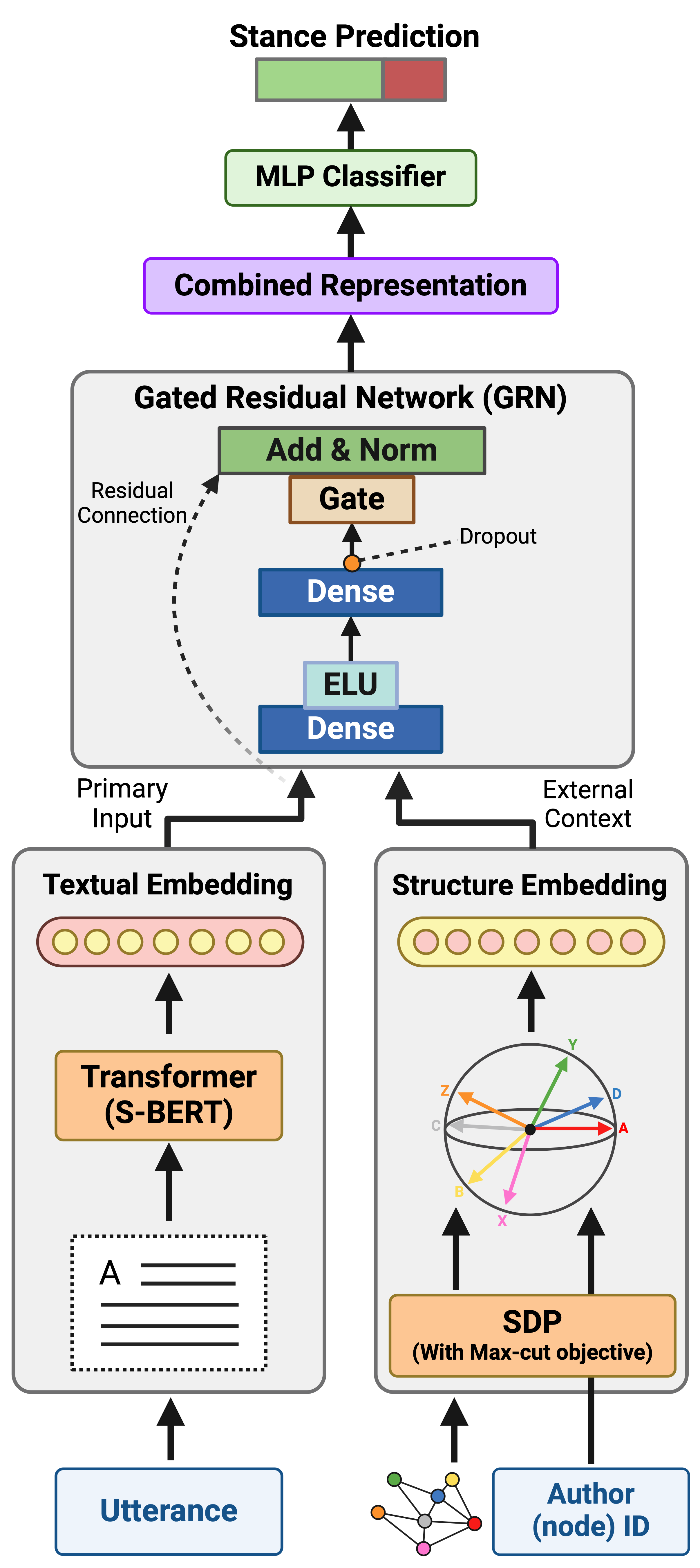}
    \caption{Illustration of the TASTE architecture.}
    \label{fig:model}
\end{figure}

The model depicted in Figure \ref{fig:model} is the utterance-level classifier. It receives as input two modalities: textual and structural. The textual modality is the embedding of the utterance, and the structural is an embedding of the speaker (details below). The two embeddings are then fused using a GRN unit, following a joint fusion strategy that dynamically combines the modalities during learning to better capture interdependencies between content and conversational structure. The fused representation is fed into an MLP classifier, which outputs a vector quantifying the likelihood of each tag $t$ in the tagset $T$ (in our case $T=\{+,-\}$), and the binary prediction is the tag with maximum likelihood.

The author's stance is computed by feeding all their utterances into the model (Figure \ref{fig:model}), getting the prediction for each utterance separately, and taking the majority vote.

In the remainder of this section, we describe the model components in more detail.


\paragraph{Textual Embedding} The representation of an utterance $u$ is derived directly from the pretrained model. Specifically, $\vec{u}$, the content embedding for $u$, is obtained as the [CLS] token vector, which is extracted by feeding $u$ to pretrained Sentence-BERT \citep{reimers2019sentencebert}.

\paragraph{Structure and User Embedding} We follow \cite{pick2022stem} in the structure representation and user embedding. The discussion tree is first converted to an interactions graph (see Figure \ref{fig:tree2graph}); each node represents a speaker, and an edge $e_{ab}$ between nodes $a$ and $b$ indicates a direct interaction between them. $w_{ab}$ indicates the weight of  $e_{ab}$, determined by the number and type of interactions. The edge weights in the interaction graph are calculated as:
\begin{align}\label{eq:InteractionEdge}
  w_{a,b} &= \alpha (replies(a, b) + replies(b, a))  \\&
  + \beta \notag (quotes(a, b) + quotes(b, a))
\end{align}
where $replies(a, b)$ and $quotes(a, b)$ indicate the number of direct replies and quotes, between users $a$ and $b$. The hyper-parameters $\alpha$ and $\beta$ are used to reflect the significance of the type of interaction, often based on the conversational norms expected in the specific platform. Through empirical testing, we determined the optimal values for these weights. For the 4Forums dataset, where interactions typically consist of direct replies to the original post (OP) and selective quoting of pertinent content, the optimal values were found to be $\alpha = 0.02$ and $\beta = 1$. Conversely, for CreateDebate, $\alpha$ was set at 1.0 and $\beta$ at 0.0, reflecting the infrequent use of quotes in this platform.


The intuition underpinning the structural embedding is that stronger disagreements lead to more intense interactions in the discussion tree, reflected as heavier-weight edges in the network. Thus, the user embeddings are trained to maximize the distance between users that are connected by heavy edges (max-cut). Mathematically, the divergence in views between $a$ and $b$ can be encoded in a distance metric $d$ between two vectors $d(\overrightarrow{a}, \overrightarrow{b}) = (1-\langle \overrightarrow{a}, \overrightarrow{b} \rangle)/2$. The maximal value of $d$ is 1 (for antipodal vectors/stances), and the minimum is 0 (when $\overrightarrow{a} = \overrightarrow{b}$, same stance). This all leads to the following optimization problem over all pairs of interacting authors:

\vspace{-17pt}
\begin{equation}\label{eq:SDP}
\mathcal{E} = \underset{\overrightarrow{a} \in S^n : \forall a \in V}{\arg\max}\sum_{(a,b)\in E} w_{ab}\frac{1- \langle \overrightarrow{a},\overrightarrow{b} \rangle}{2},
\end{equation}

where $\overrightarrow{a}$ and $\overrightarrow{b}$ are the unit vector embeddings for users $a$ and $b$ and $S^n$ is the $n$-dimensional unit sphere ($n$ is the number of users). 

The optimization program in Eq.\eqref{eq:SDP} is called a Semi-Definite Program (SDP) and can be efficiently solved using algorithms like the Ellipsoid method. We use the SDP proposed by \citet{Goemans1995} as a relaxation for the max-cut problem. Figure \ref{fig:tree2graph} illustrates the relation between the max-cut problem and the SDP solution.

\paragraph{GRN: Content and Structure Fusion} Gated Residual Networks (GRNs), as proposed by \citet{LIM20211748}, present an innovative approach to integrating a primary input vector with multiple context vectors of which relevance may vary. Notably, GRNs have proven to be particularly effective in handling datasets that are both relatively small and characterized by noise.
In its basic form, the GRN unit processes a primary input vector, denoted $a$, along with a context vector $c$:
\begin{align*}
  G&RN(a,c) = 
 LayerNorm(a + GLU(\eta_{1}),  \\&
  \eta_{1} = W_{1} \eta_{2} + b_{1}, \\ &
  \eta_{2} = ELU(W_{2} a + W_{3} c + b_{2}), \\&
  GLU(\gamma) = 
  \sigma(W_{4} \gamma + b_{4})\odot (W_{5} \gamma + b_{5})
\end{align*}
We take the user content embedding as the primary vector and the user embedding as the context, fusing them through the GRN (taking the structural embedding as the primary vector and the text embeddings as the context yielded suboptimal results).

\section{Data}
\label{sec:data}
Our analysis is conducted on two distinct datasets: 4Forums, introduced by \citet{walker2012stance}, and CreateDebate, presented by \citet{hasan-ng-2014-taking}. These datasets have been widely used in prior work on stance detection, e.g., \citet{walker2012stance, sridhar-etal-2015-joint,abbott2016internet, li2018structured, pick2022stem}. For the reader's convenience, we provide a brief overview of the datasets, emphasizing their unique characteristics. 
Descriptive statistics, comparing the two datasets are presented in Table \ref{tab:dataset_statistics}.

\begin{table}[ht] 
\centering
\resizebox{\columnwidth}{!}{
\begin{tabular}{|l|c|c|c|c|}
\hline
\textbf{Variable / Topic} & {Abortion} & {Evolution} & {Gay Marriage} & {Gun Control}\\
\hline
Num of Posts & 7,937 & 6,069 & 6,897 & 3,755\\
Num of Authors & 1,048 & 1,046 & 939 & 779\\
Num of Conversations & 51 & 53 & 49 & 49\\
Num of Interaction & 14,460 & 9,640 & 13,764 & 6,462\\
Avg. tokens per Post & 115.5 & 157.8 & 112.7 & 140.3\\
\hline
\end{tabular}
    }
    \caption*{(a) \textbf{4Forums} dataset basic statistics.}

\bigskip

\resizebox{\columnwidth}{!}{
\begin{tabular}{|l|c|c|c|c|}
\hline
\textbf{Variable / Topic} & {Abortion} & {Gay Rights} & {Marijuana} & {Obama}\\
\hline
Num of Posts & 1,741 & 1,376 & 626 & 985\\
Num of Authors & 458 & 446 & 310 & 380\\
Num of Conversations & 24 & 11 & 15 & 15\\
Num of Interaction & 1,168 & 915 & 350 & 515\\
Avg. tokens per Post & 141.9 & 111.5 & 92.9 & 109.4\\
\hline
\end{tabular}
    }
    \caption*{(b)  \textbf{CreateDebate} dataset basic statistics.}
\caption{Descriptive statistics of the datasets.}    
    \label{tab:dataset_statistics}
\end{table}

\paragraph{4Forums} 4Forums (no longer online) was a platform for political debates. Introduced by \citet{walker2012stance}, the dataset includes annotations for agree/disagree stances in 202 debates on four major topics: abortion, evolution, gay marriage, and gun control. Annotations are provided at the user level. Gold labels of utterances are derived by broadcasting the author's label to his posts.

\paragraph{CreateDebate} CreateDebate, is an online platform developed as ``a social tool that democratizes the decision-making process through online debate''.
A user (the `OP') starts a debate by posing a question (e.g., ``Should abortion be illegal: Yes or No?''). Other users can respond to the OP or to other users by adding a support, dispute, or clarification message. A benchmark containing 200 debates over four topics (abortion, gay rights, the legalization of marijuana, and Obama) was introduced by \citet{hasan-ng-2014-taking}. Self-annotation by debate participants provides gold labels at the utterance level. The most frequent self-assigned tag serves as the gold label of user.\footnote{Indeed, over 95\% of the users self-annotated all of their utterances with the same label.}

\section{Experimental Settings}
\label{sec:exp}
\subsection{Baselines} We compare our architecture to four other models based on text, structure or both:  

\paragraph{STEM and SDP} The STEM algorithm \cite{pick2022stem} is a stance classification algorithm that uses only the conversation structure. At its core, it is based on the SDP spelled in Eq.~\eqref{eq:SDP} and has shown superior results in author-level stance detection across datasets. It proceeds in three steps, the first is to compute the 2-core of the conversation graph; next compute the SDP embedding on the 2-core and derive author classifications from the vectors; finally, propagate the labels to the non-core part of the graph in a greedy manner. The derivation of author labels from SDP vector embedding is done via 
hyper-plane rounding \cite{Goemans1995}. The geometric positioning of the speaker vectors on the $n$-dimensional sphere is illustrated in Figure \ref{fig:tree2graph} (right).

A simplified version of STEM, which we call SDP, for simplicity, skips the 2-core computation and applies SDP to the entire graph. The labels are derived using the hyperplane rounding technique.

\begin{table*}[ht!]
\centering
\begin{tabular}{|l|c|c|c|c|>{\columncolor{Gray}}c|}
\hline
\textbf{Model} & Abortion & Evolution & Gay Marriage & Gun Control & Average \\
\hline
PSL \cite{sridhar-etal-2015-joint} & 77.0 & 80.3 & 80.5 & 69.1 & 76.7\\
Global \cite{li2018structured} & 86.5 & 82.2 & 88.1 & 83.1 & 84.9 \\
S-BERT \emph{(Text-Only)} & 67.7  & 69.8 & 68.9 & 71.7 & 69.5 \\
SDP \emph{(Structure-Only)}& 81.2  & 76.3 & 78.1 & 67.7 & 75.8 \\
\hline
\textbf{\emph{$\mathbf{TASTE}@\scriptstyle{GRN(N2V+SBERT)}$}} & 67.9  & 70.2 & 69.0 & 71.1 & 69.6 \\
\textbf{\emph{$\mathbf{TASTE}@\scriptstyle{CNCT(SDP+SBERT)}$}} & 83.0  & 80.5 & 80.2 & 76.2 & 79.9 \\
\textbf{\emph{$\mathbf{TASTE}@\scriptstyle{GRN(SDP+SBERT)}$}} & \textbf{90.5} & \textbf{88.9}& \textbf{88.4} & \textbf{83.2} & \textbf{87.7} \\
\hline
\end{tabular}
\caption*{(a) \textbf{Post} stance classification accuracy for \textbf{4Forums} dataset.}

\bigskip 

\begin{tabular}{|l|c|c|c|c|>{\columncolor{Gray}}c|}
\hline
\textbf{Model} & Abortion & Evolution & Gay Marriage & Gun Control & Average \\
\hline
PSL \cite{sridhar-etal-2015-joint} & 66.0 & 78.7 & 77.1 & 68.3 & 72.5 \\
STEM \cite{pick2022stem} & 78.2 & 75.5 & 77.0 & 71.2 & 75.6 \\
S-BERT \emph{(Text-Only)}& 71.1 & 79.5 & 78.0 & 76.8 & 76.3 \\
SDP \emph{(Structure-Only)}& 71.6  & 81.9 & 80.1 & 68.1 & 75.4 \\
\hline
\textbf{\emph{$\mathbf{TASTE}@\scriptstyle{GRN(N2V+SBERT)}$}} & 70.3  & 79.9 & 79.2 & 76.2 & 76.4 \\
\textbf{\emph{$\mathbf{TASTE}@\scriptstyle{CNCT(SDP+SBERT)}$}} & 74.1  & 82.7 & 81.0 & 77.5 & 78.8 \\
\textbf{\emph{$\mathbf{TASTE}@\scriptstyle{GRN(SDP+SBERT)}$}} & \textbf{79.6} & \textbf{83.9} & \textbf{82.5} & \textbf{78.8} & \textbf{81.2} \\
\hline
\end{tabular}
\caption*{(b) \textbf{Author} stance classification accuracy for \textbf{4Forums} dataset.}

\bigskip 

\begin{tabular}{|l|c|c|c|c|>{\columncolor{Gray}}c|}
\hline
\textbf{Model} & Abortion & Gay Rights & Marijuana & Obama & Average \\
\hline
PSL \cite{sridhar-etal-2015-joint} & 66.8 & 72.7 & 69.1 & 63.7 & 68.0 \\
Global \cite{li2018structured} & \textbf{81.1} & 77.2 & 77.6 & 64.8 & 75.1 \\
S-BERT \emph{(Text-Only)} & 63.0 & 64.8 & 71.9 & 58.9 & 64.6 \\
SDP \emph{(Structure-Only)}& 71.4  & 69.8 & 70.6 & 57.0 & 67.2 \\
\hline
\textbf{\emph{$\mathbf{TASTE}@\scriptstyle{GRN(N2V+SBERT)}$}} & 62.8  & 65.0 & 71.8 & 60.2 & 64.9 \\
\textbf{\emph{$\mathbf{TASTE}@\scriptstyle{CNCT(SDP+SBERT)}$}} & 71.5  & 78.8 & 72.2 & 70.9 & 73.3 \\
\textbf{\emph{$\mathbf{TASTE}@\scriptstyle{GRN(SDP+SBERT)}$}} & 80.1 & \textbf{82.5} & \textbf{78.0} & \textbf{78.7} & \textbf{79.8} \\
\hline
\end{tabular}
\caption*{(c) \textbf{Post} stance classification accuracy for \textbf{CreateDebate} dataset.}

\bigskip 

\begin{tabular}{|l|c|c|c|c|>{\columncolor{Gray}}c|}
\hline
\textbf{Model} & Abortion & Gay Rights & Marijuana & Obama & Average \\
\hline
PSL \cite{sridhar-etal-2015-joint} & 69.5 & 74.0 & 75.4 & 66.1 & 71.2 \\
STEM \cite{pick2022stem}& \textbf{85.7} & 79.9 & 70.5 & \textbf{82.3} & \textbf{79.6} \\
S-BERT \emph{(Text-Only)} & 63.9 & 70.3 & 75.3 & 55.3 & 66.2 \\
SDP \emph{(Structure-Only)}& 61.0  & 75.5 & 76.3 & 56.0 & 67.2 \\
\hline
\textbf{\emph{$\mathbf{TASTE}@\scriptstyle{GRN(N2V+SBERT)}$}} & 64.8  & 70.1 & 75.6 & 56.2 & 66.6 \\
\textbf{\emph{$\mathbf{TASTE}@\scriptstyle{CNCT(SDP+SBERT)}$}} & 67.6  & 79.0 & 78.0 & 62.6 & 71.8 \\
\textbf{\emph{$\mathbf{TASTE}@\scriptstyle{GRN(SDP+SBERT)}$}} & 71.9 & \textbf{81.7} & \textbf{79.6} & 74.9 & 77.0 \\
\hline
\end{tabular}
\caption*{(d) \textbf{Author} stance classification accuracy for \textbf{CreateDebate} dataset.}

\caption{Average accuracy on authors’ and posts' stance classification of 4Forums and CreateDebate datasets across eight discussion topics. $\mathbf{TASTE}$ architecture differ in their fusion method and the type of structural embedding used: Gated Residual Network ($\scriptstyle{GRN}$) and concatenation ($\scriptstyle{CNCT}$) techniques for fusion, alongside employing either Node2Vec ($\scriptstyle{N2V}$) or Semi-Definite Programming ($\scriptstyle{SDP}$) for structural embeddings. Highest scores for each topic are highlighted in bold.}
\label{tab:comprehensive_res}
\end{table*}

\paragraph{PSL} 
The Probabilistic Soft Logic (PSL) \cite{sridhar-etal-2015-joint} approach combines the expressiveness of first-order logic with the probabilistic modeling capabilities of graphical models. It allows for the flexible specification of complex, relational structures and dependencies in the data, making it well-suited for tasks such as stance detection. PSL formulates the problem as a joint probabilistic inference task, where the goal is to infer the most likely values of the unobserved variables (such as stance labels) given the observed data (such as textual features).

\paragraph{S-BERT} Sentence-BERT is a modification of the BERT model specifically designed for sentence embeddings. Developed by \cite{reimers2019sentencebert}, S-BERT fine-tunes BERT by training it on a siamese and triplet network architecture, which allows it to learn better sentence embeddings. These embeddings are capable of capturing semantic similarity between sentences, making them useful for various natural language processing tasks like sentence classification, semantic search, and clustering. S-BERT has been shown to outperform traditional BERT embeddings in tasks that require an understanding of sentence-level semantics.

\paragraph{Global} The Global Embedding model, as introduced by \citet{li2018structured}, leverages both text and structural information within online debates to create a unified global embedding. Unlike traditional methods that might treat text and structure separately, their method captures the nuanced interplay between an author's contributions and the broader conversational context. While their approach closely aligns with our methodology in considering both structural and textual information, their model integrates these two dimensions into a single global embedding, contrasting with our technique of generating distinct embeddings for text and structure separately.

\subsection{Technical Specifications} We trained TASTE with a maximum of 10 epochs, employing the AdamW optimizer \cite{loshchilov2019decoupled} with a batch size of 16. A learning rate decay strategy was utilized, starting the learning rate within the range of $10^{-5}$. This rate was halved each time the validation loss showed no improvement every three epochs. The training was terminated when either the learning rate was reduced to the minimum threshold of $10^{-8}$, or when the maximum epoch limit of 10 was reached. 
To avoid data leakage, we ensured that posts from the same author were not included in both training and test sets simultaneously. 
In both training and testing we used Google's co-lab environment with T4 GPU. Training each TASTE version, and also running each experiment ran for on average for no more than two hours.


\section{Results and Analysis}
\label{sec:res}
Table \ref{tab:comprehensive_res}(a--d) compares TASTE and the baseline models for utterance and user level over the two benchmarks and across topics. The table also provides results of three variations of the TASTE architecture: (1) $\mathbf{TASTE}@\scriptstyle{GRN(SDP+SBERT)}$ uses GRN to fuse the SDP and the S-BERT embedding; (2) $\mathbf{TASTE}@\scriptstyle{CNCT(SDP+SBERT)}$ skips the GRN, concatenates the two embeddings and pushes them into the MLP layer; (3) $\mathbf{TASTE}@\scriptstyle{GRN(N2V+SBERT)}$ uses a node2vec \cite{grover2016node2vec} instead of SDP.

Keeping in line with previous works, we use accuracy as the base metric to evaluate the models. Reported values are average accuracy of a 5-fold cross-validation setting. All results achieved statistical significance with $p$-values less than $0.05$ in paired $t$-tests comparing the $\mathbf{TASTE}$ model against the baseline models.

$\mathbf{TASTE}@\scriptstyle{GRN(SDP+SBERT)}$ achieves best performance in 3 out of 4 tasks (\ref{tab:comprehensive_res}a-c), across all topics.  $\mathbf{TASTE}@\scriptstyle{CNCT(SDP+SBERT)}$ was ranked second, trailing STEM, in the fourth task at Table \ref{tab:comprehensive_res}d. However, replicating the work of \citet{pick2022stem} we observed that they excluded users lacking strong inter-annotator agreement; these are probably the more challenging cases. In contrast, our application of TASTE and the SDP considers all users, hence the observed differences in performance in some cases.
The other two versions of TASTE are often competitive but are outperformed by some of the baselines. This trend emphasizes the advantage provided by the learning SDP embeddings, compared to node2vec, and of the use of the GRN unit for vectors fusion, compared to concatenation.
Finally, we observe that SDP alone achieves excellent results -- being the strongest of the baselines in tasks a--c.

\paragraph{Content vs. Structure} Using uni-modal approach, we can tease the content and the structure apart and gain some interesting insights about interplay between them. Intense (dense) though short (in utterance length) interactions, such as those in the Abortion, Evolution, and Gay Marriage topics, are better captured by structural models such as the SDP, while less intense interactions that exhibit longer utterances, such as in Gun Control, are better captured by text based embeddings such as S-BERT. The uni-modal results along with the number of interactions and the number of tokens in each topic are provided in Table \ref{tab:res2}. 

\begin{table}[ht!] 
\centering
\resizebox{\columnwidth}{!}{
    \begin{tabular}{|l|c|c|c|c|}
        \hline
 & Abortion & Evolution  & Gay Marriage & Gun Control \\
\hline
\# Interaction &  14,460 & 9,640 &  13,764 & 6,462\\
Avg. tokens & 115.5 & 157.8 & 112.7 & 140.3\\
\hline
S-BERT (Acc.) & 67.7 & 69.8 & 68.9 & \textbf{71.7}\\
SDP (Acc.) & \textbf{81.2} & \textbf{76.3} & \textbf{78.1} & 67.7\\ 
\hline
    \end{tabular}
    }
    \caption{Post stance classification on the 4Forums dataset comparing pure textual vs. pure structural approaches (S-BERT vs SDP). The number of interactions and the average length, per topic, are listed.}
    \label{tab:res2}
\end{table}


This pattern can be explained in a number of ways: (i) in highly interactive but less verbose environments, the structure of interactions becomes more important in indicating stance. The structure of the discourse, encompassing aspects such as the frequency and network of replies, becomes a potent indicator of the users' stances, and (ii) long argumentation is correlated with sparser networks since participation is more demanding. On the other hand -- the longer texts provide stronger signal to be exploited by text-based models.

These results suggest that the effectiveness of structure vs. content-focused models also depends on the conversational dynamics that constitute the (local) social context.

\vspace{1.5mm}

Integrating content and structural information in a multimodal manner proves robust across datasets, topics, and conversational dynamics. The multimodal approach grounds the potentially rich textual content in the relevant social context. 
\begin{figure}[ht!]
    \centering
    \includegraphics[width=\columnwidth, height=13cm]{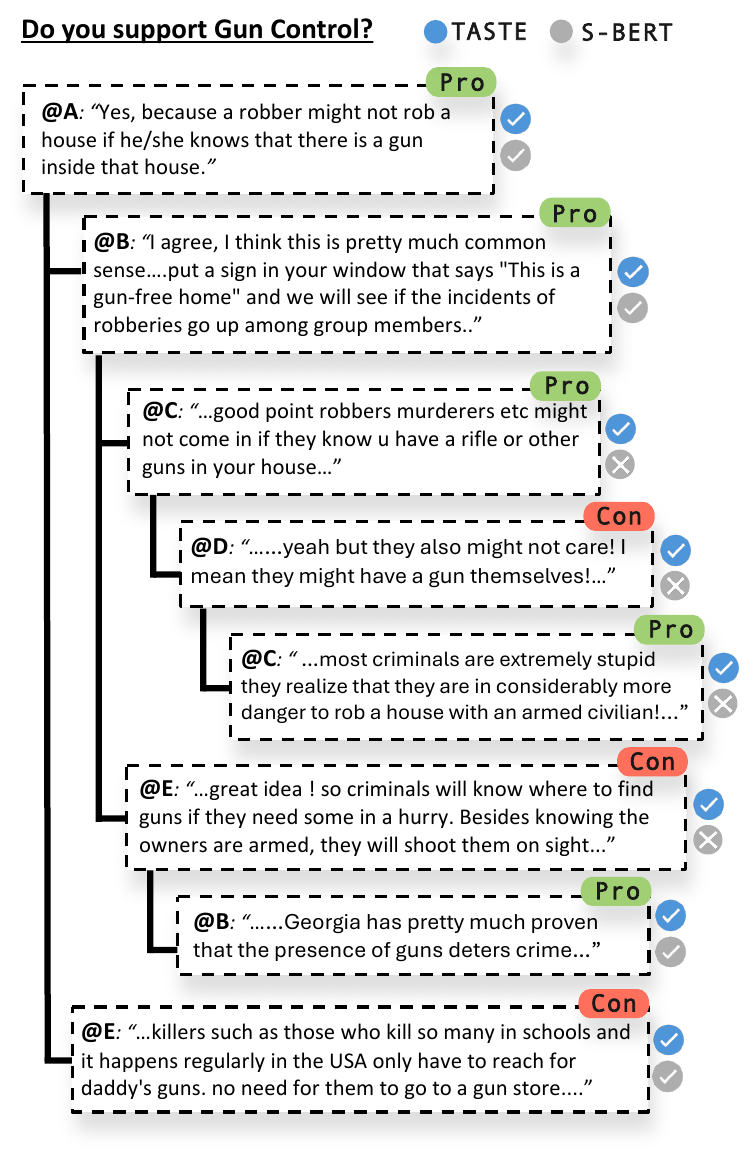}
    \caption{A short excerpt from a longer discussion (“Do you support Gun Control?”) in 4Forums. Pro/Con indicates the true label. \cmark  or \xmark~ in a blue (gray) circle indicate whether TASTE (S-BERT) assigned the correct label to the utterance.}
    \label{fig:4forums_tree}
\end{figure}

The exchange presented in Figure \ref{fig:4forums_tree} provides an example for the value of the multimodal framework and the contribution of the SDP to understanding the global structure of a discussion. In this discussion on gun control, participants engage in a more complex interaction pattern than simple turn-taking, where each comment directly opposes the previous one. Instead, participants often respond supportively to others’ comments, creating a nuanced dynamic which locally violates the max-cut assumption, a violation that may also affect the tone of the comment. The first comment of C, which locally violates the max-cut hypothesis, is such an example of supportive utterance which reads quite ambiguous when taken out of context. However, by adding the global overview of the SDP to the model, it can accurately capture these interactions and correctly classify the participants’ stances, circumventing the local max-cut violation and the diluted tone of the utterance. Diving into specific results, relying solely on SBERT led to a 71\% accuracy for the entire conversation from which this excerpt is taken, while our best TASTE model achieved a 92\% accuracy. This significant improvement underscores the effectiveness of combining textual and structural embeddings for nuanced stance detection.


\paragraph{Node Embeddings: SDP vs. node2vec} Using SDP to obtain node embeddings proved superior to the more traditional node2vec approach, with performance gains between 0.09 and  0.17. We attribute the difference in performance to two factors: (i) node2vec performs best on networks much larger than the conversational networks in our datasets, and (ii) The node2vec approache aims at minimizing the distances between neighbouring nodes, while in our case we actually try to separate them, given the intuition that they reflect opposite stances. The SDP, on the other hand is designed to achieve exactly that via the max-cut.

\paragraph{Fusion Strategy}
We considered a naive yet commonly used fusion strategy, namely concatenation of the representations (see results in Table \ref{tab:res3} in Appendix \ref{sec:appendix_fusion}). Using a GRN layer significantly outperforms concatenation across all topics. This notable difference in the results can be attributed to the GRN layer's ability to scrupulously combine the distinct properties of both content and structure embeddings. Unlike simple concatenation, which merely amalgamates or adds up the information, the GRN layer applies a more nuanced approach. It effectively `gates' the information from each embedding, allowing for a selective integration process. This gating mechanism is particularly advantageous in handling smaller and noisier datasets where discerning relevant information is critical. The GRN layer's capacity to attend to the the most pertinent features from both content and structural data results in a more accurate and robust stance detection model.


\paragraph{Stance and the Number of Utterances} Examining the error rate of TASTE at the author level with regards to the number of utterances per user, we observed that the error rate over users with only one or two utterances is $\times 1.5$ the average error rate, and an order of magnitude higher than the error rate over users with twenty utterances or more (which drops to only 0.05). Nevertheless, TASTE outperforms other models even over the users with a low number of utterances. These results again confirm our intuition: Sharp stance differences are reflected by the conversation dynamics. The more intense the conversation -- the more utterances a user makes (high engagement), providing a stronger signal in both modalities. However, the structural embeddings enhance the signal even in cases of a curtailed contribution of one or two utterances. 

The performance gap between TASTE and other models is notably smaller at the author level than at the post level. This can be explained by the inference stage methodology, which determines a user's stance through majority voting across all their posts. Consequently, if a model accurately predicts the majority stance of a user's posts, it achieves perfect accuracy for that user, which is a harder task than correctly classifying each individual post by that same user. Therefore, the latter task is more effective at highlighting performance differences between models.

\section{Discussion}
\label{sec:discussion}

\paragraph{Social Grounding and Multimodality}
In linguistics, cognitive science, and communication studies the concept of \emph{grounding} refers to a communication phase in which the interlocutors assume or establish the common ground required for mutual understanding \cite{harnad1990symbol,clark1991grounding,lewis2008convention}. In the field of AI (NLP, Autonomous Agents, etc.), the term `grounding' usually means that models are trained with respect to other modalities that reflect the ``environment''. Typical examples include visual and audio modalities, e.g.,  \cite{ngiam2011multimodal,mao2016training,yao-mihalcea-2022-modality}. This latter practical approach to grounding could be viewed as a limited implementation of the broader concept of communication grounding \cite{chandu2021grounding}. Our results empirically confirm fundamental frameworks such as face work \cite{goffman1955face} and the cooperative principle \cite{grice1975logic} -- in which texts are interpreted upon grounding in the (limited) social context -- the conversation graph. 

\paragraph{Datasets and Conversational Dynamics} Some recent work (see Section \ref{sec:rw}) use other, more recent, datasets. The focus of this work is the interplay between content and speaker dynamics, hence these datasets cannot be used for proper comparison of the utterance level, nor for speaker-level stance. 

\paragraph{Modalities, Fusion strategies (and terminology)}
The work of \citet{li2018structured}, explicitly modeling text and structure, is the most similar to our work. However, while their strategy for integrating text, speaker, and structure can be viewed as early fusion\footnote{We note that Li et al. do not refer to their model as multimodal and do not explicitly refer to the fusion strategy.}, where modalities are combined at the input level to produce global embeddings in one shared space, our methodology combines embeddings of the different modalities using joint (hierarchical) fusion strategy \cite{huang2020fusion,multimodalTransformersSurvey2023}, where interactions between modalities are dynamically modeled during the learning process. We believe that the joint fusion of separate embeddings provides flexibility that results in higher efficiency. Specifically, it allows the use of the SDP for speaker representation. These non-orthodox embeddings are learned efficiently at the conversation level in a completely unsupervised manner.

\section{Conclusion}
\label{sec:con}

We introduced TASTE, a multimodal model fusing Text and STurcture Embedding model, designed for stance detection. The model effectively leverages the intricate interplay of conversation content and structure to compute comprehensive user embeddings. We have shown that the application of a GRN layer, initially utilized for multi-horizon time-series forecasting as described by \citet{LIM20211748}, is particularly advantageous for the task we are addressing, especially considering the constraints of our relatively small dataset. Our evaluation of the 4Forum and CreateDebate datasets, alongside comparisons with state-of-the-art models, highlights the distinct advantages of TASTE.

Furthermore, our analysis revealed a notable correlation between stance classification accuracy and the balance of textual depth and interaction frequency. Specifically, in scenarios where 
participant interactions are frequent yet text contributions are succinct, structural-based models gain an upper hand. Conversely, in conversations characterized by lengthier texts but fewer interactions, content-based models excel. 

While uni-modal approaches have their respective advantages in certain scenarios, it is the integration of both these elements that consistently leads to the most solid and reliable outcomes.
This combined approach not only delves into the detailed and subtle aspects of the textual content but also leverages the patterns and complexities in human interactions, in line with theoretical frameworks established by \cite{goffman1955face,goffman1978response,grice1975logic} and others. 

\section{Limitations}
Our approach suffers from a number of limitations that would be addressed in future work.  

First, although the 4Forums and CreateDebate datasets are commonly used and allow us to compare our work to relevant previous works, further evaluation on more diverse and contemporary datasets, such as social media platforms like X (Twitter) or Reddit, would improve the generalizability of our findings. 

A second, though related, limitation stems from the focus on English datasets. The performance of the model highly depends on conversational norms and dynamics, which may vary across languages and platforms.

Finally, the model, at least when applied on the user level, is based on the implicit assumption that a speaker holds a pre-formed and stable stance throughout the discussion. While this assumption holds in the datasets we use, it may not hold in other datasets, e.g., Reddit's Change My View in which users are encouraged to persuade each other: ``A place to post an opinion you accept may be flawed, in an effort to understand other perspectives on the issue. Enter with a mindset for conversation, not debate.'' Future work should address the (optimistic) case in which speakers are less dogmatic and open to change of hearts. We plan to address this by combining our architecture that depends on the global conversation structure with the local turn taking recently explored by \cite{wen2024transitive}.

\bibliography{custom}
\newpage
\appendix

\section{GRN and Naive Fusion}
\label{sec:appendix_fusion}

\begin{table}[h!] 
\centering
\resizebox{\columnwidth}{!}{
    \begin{tabular}{|l|c|c|c|c|}
        \hline
        \multicolumn{5}{|c|}{\cellcolor[gray]{0.9} 4Forums Dataset} \\
        \hline
        & Abortion & Evolution  & Gay Marriage & Gun Control \\
        \hline       \textbf{\emph{$\mathbf{TASTE}@\scriptstyle{CNCT}$}} (Acc.) &  83.0 & 80.5 &  80.2 & 76.2 \\ 
        \hline \textbf{\emph{$\mathbf{TASTE}@\scriptstyle{GRN}$}} (Acc.) &  \textbf{90.5} & \textbf{88.9} &  \textbf{88.4} & \textbf{83.2} \\
        \hline
        \multicolumn{5}{|c|}{\cellcolor[gray]{0.9}CreateDebate Dataset} \\
        \hline
        & Abortion & Gay Rights  & Marijuana & Obama \\
        \hline
\textbf{\emph{$\mathbf{TASTE}@\scriptstyle{CNCT}$}} (Acc.) &  71.5 & 78.8 &  72.2 & 70.9 \\ 
        \hline \textbf{\emph{$\mathbf{TASTE}@\scriptstyle{GRN}$}} (Acc.) &  \textbf{80.1} & \textbf{82.5} &  \textbf{78.0} & \textbf{78.7} \\
        \hline
    \end{tabular}
    }
    \caption{Post stance classification accuracy comparison for CD and 4Forums datasets. The GRN version consistently outperforms the CNCT version across all topics, illustrating the effectiveness of integrating content and structure through the GRN layer.}
    \label{tab:res3}
\end{table}

Table \ref{tab:res3} provides the GRN and the the concatenations results for the two datasets.


\end{document}